\documentclass[12pt]{article}

\usepackage[utf8]{inputenc}
\usepackage[T1]{fontenc}
\usepackage{graphicx}
\usepackage{booktabs}
\usepackage{amsmath}
\usepackage{hyperref}
\usepackage[numbers]{natbib}
\usepackage{geometry}
\usepackage{array}
\usepackage{makecell}  

\title{\textbf{Performance of a Deep Learning-Based Segmentation Model for Pancreatic Tumors on Public Endoscopic Ultrasound Datasets}}

\author{}
\date{}

\begin{document}
\maketitle

\vspace{-4em}

\begin{center}
Pankaj Gupta\textsuperscript{1,*}, Priya Mudgil\textsuperscript{1}, Niharika Dutta\textsuperscript{1}, Kartik Bose\textsuperscript{1},\\
Nitish Kumar\textsuperscript{1}, Anupam Kumar\textsuperscript{2}, Jimil Shah\textsuperscript{2}, Vaneet Jearth\textsuperscript{2},\\
Jayanta Samanta\textsuperscript{2}, Vishal Sharma\textsuperscript{2}, Harshal Mandavdhare\textsuperscript{2},\\
Surinder Rana\textsuperscript{2}, Saroj K Sinha\textsuperscript{2}, Usha Dutta\textsuperscript{2}
\end{center}

\footnotetext[1]{Department of Radiodiagnosis, Postgraduate Institute of Medical Education and Research, Chandigarh, India 160012}
\footnotetext[2]{Department of Medical Gastroenterology, Postgraduate Institute of Medical Education and Research, Chandigarh, India 160012}
{\renewcommand{\thefootnote}{\fnsymbol{footnote}}\footnotetext[1]{Corresponding Author: pankajgupta959@gmail.com}}

\vspace{2em}
\begin{abstract}
\noindent\textbf{Background:} Pancreatic cancer is one of the most aggressive cancers, with poor survival rates. Endoscopic ultrasound (EUS) is a key diagnostic modality, but its effectiveness is constrained by operator subjectivity. This study evaluates a Vision Transformer-based deep learning segmentation model for pancreatic tumors.

\vspace{0.5em}
\noindent\textbf{Methods:} A segmentation model using the USFM framework with a Vision Transformer backbone was trained and validated with 17,367 EUS images (from two public datasets) in 5-fold cross-validation. The model was tested on an independent dataset of 350 EUS images from another public dataset, manually segmented by radiologists. Preprocessing included grayscale conversion, cropping, and resizing to 512$\times$512 pixels. Metrics included Dice similarity coefficient (DSC), intersection over union (IoU), sensitivity, specificity, and accuracy.

\vspace{0.5em}
\noindent\textbf{Results:} In 5-fold cross-validation, the model achieved a mean DSC of 0.651 $\pm$ 0.738, IoU of 0.579 $\pm$ 0.658, sensitivity of 69.8\%, specificity of 98.8\%, and accuracy of 97.5\%. For the external validation set, the model achieved a DSC of 0.657 (95\% CI: 0.634--0.769), IoU of 0.614 (95\% CI: 0.590--0.689), sensitivity of 71.8\%, and specificity of 97.7\%. Results were consistent, but 9.7\% of cases exhibited erroneous multiple predictions.

\vspace{0.5em}
\noindent\textbf{Conclusions:} The Vision Transformer-based model demonstrated strong performance for pancreatic tumor segmentation in EUS images. However, dataset heterogeneity and limited external validation highlight the need for further refinement, standardization, and prospective studies.
\end{abstract}

\vspace{1em}

\section{Introduction}

Pancreatic cancer remains a highly lethal malignancy, with a global five-year survival rate of less than 10\% \cite{ref1}. The disease's asymptomatic nature in early stages results in most patients being diagnosed at advanced stages, emphasizing the critical need for early and accurate detection. Conventional imaging modalities such as computed tomography and magnetic resonance imaging lack sensitivity in detecting smaller lesions and differentiating benign from malignant masses \cite{ref2, ref3}. Endoscopic ultrasound (EUS) offers higher sensitivity and spatial resolution, enabling the visualization of smaller pancreatic abnormalities \cite{ref4}. EUS also facilitates procedures such as fine-needle aspiration (FNA) for tissue diagnosis, enhancing diagnostic accuracy \cite{ref2}.

However, EUS interpretation heavily relies on operator expertise. Factors such as inter-observer variability, and learning curve can lead to inconsistent interpretations and diagnostic errors \cite{ref5}. Recent advancements in artificial intelligence (AI) allow integrating deep learning (DL) techniques into medical imaging workflows \cite{ref6}. DL-based segmentation models provide a means to automate lesion detection and segmentation, thereby minimizing operator dependency and standardizing diagnostic outcomes \cite{ref7}.

While convolutional neural networks (CNNs) have been widely adopted for tumor segmentation in EUS, Vision Transformer (ViT) models have recently emerged as an alternative \cite{ref8}. ViT models excel in capturing long-range dependencies and spatial relationships, often outperforming traditional architectures in segmentation tasks. This study evaluates a Vision Transformer-based segmentation model trained on publicly available EUS datasets and tested on an external public dataset. The study aims to validate the model's segmentation performance and assess its generalizability across heterogeneous datasets.

\section{Methods}

\subsection{Dataset Description}

\subsubsection{Training Dataset}
The training dataset included 17,367 EUS images from two publicly available sources:

\begin{enumerate}
    \item \textbf{Pancreatic Cancer Dataset} \cite{ref9}: The pancreatic cancer dataset comprised 18 cases, representing 16,853 frames extracted from EUS video sequences. The patients had a mean age of 65.2 years (range: 50--87 years) and included 10 males and 8 females. Tumors were predominantly located in the head of the pancreas, with fewer cases involving the pancreatic body and tail. Tumor sizes ranged from 15 mm to 43.8 mm, with an average size of approximately 32.9 mm. TNM staging, where reported, included 2 cases of T1, 1 case of T2, 11 cases of T3, and 2 cases of T4 tumors. Nodal involvement (N-stage) was reported as N0 in 8 cases and N1 in 5 cases, while 3 cases had indeterminate nodal status (NX). Evidence of distant metastases was present in 6 cases, with the remaining cases having unreported metastasis status (MX).
    
    \item \textbf{GIST514-DB Dataset} \cite{ref10}: The GIST514-DB included 514 EUS images, with 263 GISTs and 251 leiomyomas. Patients with GISTs had a mean age of 59.9 $\pm$ 8.7 years, compared to 54.5 $\pm$ 10.3 years for those with leiomyomas, with no significant gender differences between groups. GISTs were predominantly located in the fundus (202 cases) and body (41 cases), whereas leiomyomas were mostly in the esophagus (128 cases) and cardia (18 cases). GISTs had a mean horizontal diameter of 10.9 $\pm$ 5.8 mm, compared to 10.1 $\pm$ 6.0 mm for leiomyomas, with longitudinal dimensions significantly larger in GISTs (7.5 $\pm$ 4.5 mm vs. 6.2 $\pm$ 3.6 mm; $p < 0.001$). Tumor risk stratification revealed a predominance of very low-risk GISTs (218 cases, 82.9\%), with a minority categorized as low, intermediate, or high risk. This dataset provided comprehensive segmentation annotations, enabling use for training in lesion segmentation tasks.
\end{enumerate}

\subsubsection{Testing Dataset}
\textbf{LEP Dataset:} External validation was performed on 350 hand-curated EUS images from the LEP dataset \cite{ref11}. The LEP dataset is a large-scale repository of EUS-based images collected by the Department of Gastroenterology, Changhai Hospital, Second Military Medical University/Naval Medical University. The labelled subset of the dataset contains 3,500 EUS images divided into two categories: pancreatic cancer (PC; 1,680 images) and non-pancreatic cancer (NPC; 1,820 images). Images were sourced from 420 patients (280 from pancreatic cancer; 140 from NPC). For external testing, 350 pancreatic cancer images were selected from the 1,680 labeled images. Inclusion criteria focused on high-quality images with clear tumor representation and lesion absence of Doppler artifacts.

\subsection{Preprocessing}
All images underwent preprocessing using consistent protocols. Metadata around the periphery was cropped, and images were resized to 512$\times$512 pixels using bicubic interpolation. The frames were converted to grayscale, with no additional normalization or augmentation applied.

\subsection{Model Architecture}
The segmentation model was implemented using the USFM framework \cite{ref12} with a Vision Transformer backbone, HVITBackbone4Seg. The backbone divided input images into 16$\times$16 patches, with an embedding dimension of 768 and 12 layers of depth. Relative positional biases were employed instead of absolute encodings. Segmentation was driven by an ATMHead decoder, using three layers and 12 attention heads. The ATMLoss function computed binary masks (foreground vs. background). The complete model architecture and training hyperparameters are presented in Table~\ref{tab:model-arch}.

The model was trained for 50 epochs using 5-fold cross-validation. Training utilized the AdamW optimizer with a cosine learning rate schedule. Validation occurred every five epochs, with the best-performing checkpoint identified based on the highest Dice score. For testing, the model predicted binary segmentation masks from logits without connected-component filtering or post-processing.

Metrics used for evaluation included Dice similarity coefficient (DSC), intersection over union (IoU), sensitivity, specificity, and accuracy. Ninety-five percent confidence intervals (95\% CI) were computed. Additionally, a failure analysis was done.

As the study exclusively utilized data obtained from a publicly available dataset with no access to identifiable or sensitive patient information, formal ethical approval was not required for the conduct of this research.

\begin{table}[t!]
\centering
\small
\caption{Model Architecture and Training Hyperparameters}
\label{tab:model-arch}
\renewcommand{\arraystretch}{1.2}
\begin{tabular}{|p{4cm}|p{11cm}|}
\hline
\textbf{Parameter} & \textbf{Details} \\
\hline
Model Framework & USFM-based segmentation model utilizing Vision Transformer (HVITBackbone4Seg) \\ \hline
Backbone & Vision Transformer \\ \hline
Patch Size & 16$\times$16 \\ \hline
Embedding Dimension & 768 \\ \hline
Depth & 12 layers \\ \hline
Attention Heads & 12 \\ \hline
Relative Positional Bias & Used (No absolute positional encoding) \\ \hline
Feature Taps & Block outputs from layers 5, 7, and 11 \\ \hline
Decoder (ATMHead) & \makecell[l]{
    Input dimension: 512$\times$512\\
    Channels: 768\\
    Embedding Dimension: 384\\
    Layers: 3\\
    Attention Heads: 12\\
    Loss Function: ATMLoss (num\_classes = 2, dec\_layers = 3)
} \\ \hline
Training Dataset & 17,367 EUS images (Pancreatic Cancer + GIST) \\ \hline
Testing Dataset & 350 curated EUS images from the LEP dataset \\ \hline
Optimizer & AdamW \\ \hline
Learning Rate & Initial: 3$\times$10$^{-4}$, Warmup (20 epochs) from 5$\times$10$^{-5}$, Cosine decay \\ \hline
Weight Decay & 0.05 \\ \hline
Layer Decay & 0.65 \\ \hline
Gradient Clipping & 5.0 \\ \hline
Batch Size & 16 (Global Batch Size, Mixed Precision FP16) \\ \hline
Training Epochs & 50 \\ \hline
Validation Frequency & Every 5 epochs, using held-out fold metrics \\ \hline
Best Checkpoint & Highest validation Dice similarity coefficient \\ \hline
Inference Method & Argmax over logits to obtain binary masks \\ \hline
Hardware Setup & 2 GPUs (NVIDIA RTX 6000 ADA (48 GB), CUDA-enabled \\
\hline
\end{tabular}
\end{table}

\section{Results}

During cross-validation, the model demonstrated consistent performance across the training and validation datasets. The mean DSC was 0.658 [95\% confidence interval (CI) 0.615--0.738] and IoU score was 0.579 (95\% CI 0.557--0.658). Specificity was high, averaging 98.8\%, whereas the sensitivity was 69.8\%. The overall accuracy of the model across folds was 97.5\%.

\begin{figure}[htbp]
\centering
\includegraphics[width=\textwidth]{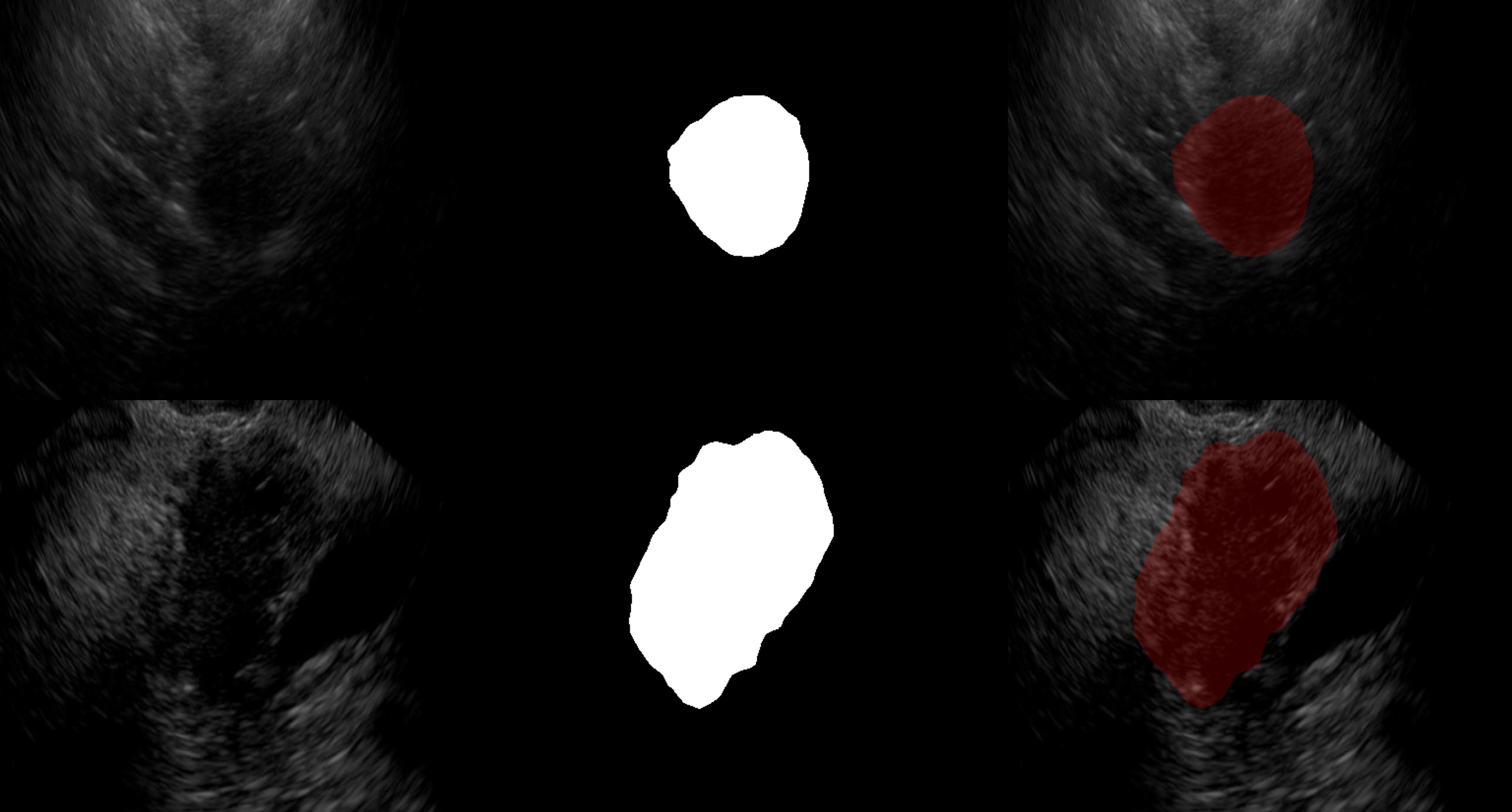}
\caption{Segmentation visualization of pancreatic cancer in 2 patients with excellent segmentation by the DL model. Each row shows (left) input EUS image, (middle) ground-truth mask, (right) model prediction overlay (red). The DSC of the 1st patient (upper row) was 0.891 and that of the 2nd patient (lower row) was 0.905.}
\label{fig:excellent-segmentation}
\end{figure}

On the external test dataset of 350 images, the model achieved a DSC of 0.657 (95\% CI 0.634--0.769) (Figure~\ref{fig:excellent-segmentation}). The IoU for the test set was 0.614 (95\% CI: 0.590--0.689). Sensitivity on this dataset was 71.8\% (95\% CI 69.1--79.3), while specificity was 97.7\% (95\% CI 95.1--99.2).

\begin{figure}[t!]
\centering
\includegraphics[width=\textwidth]{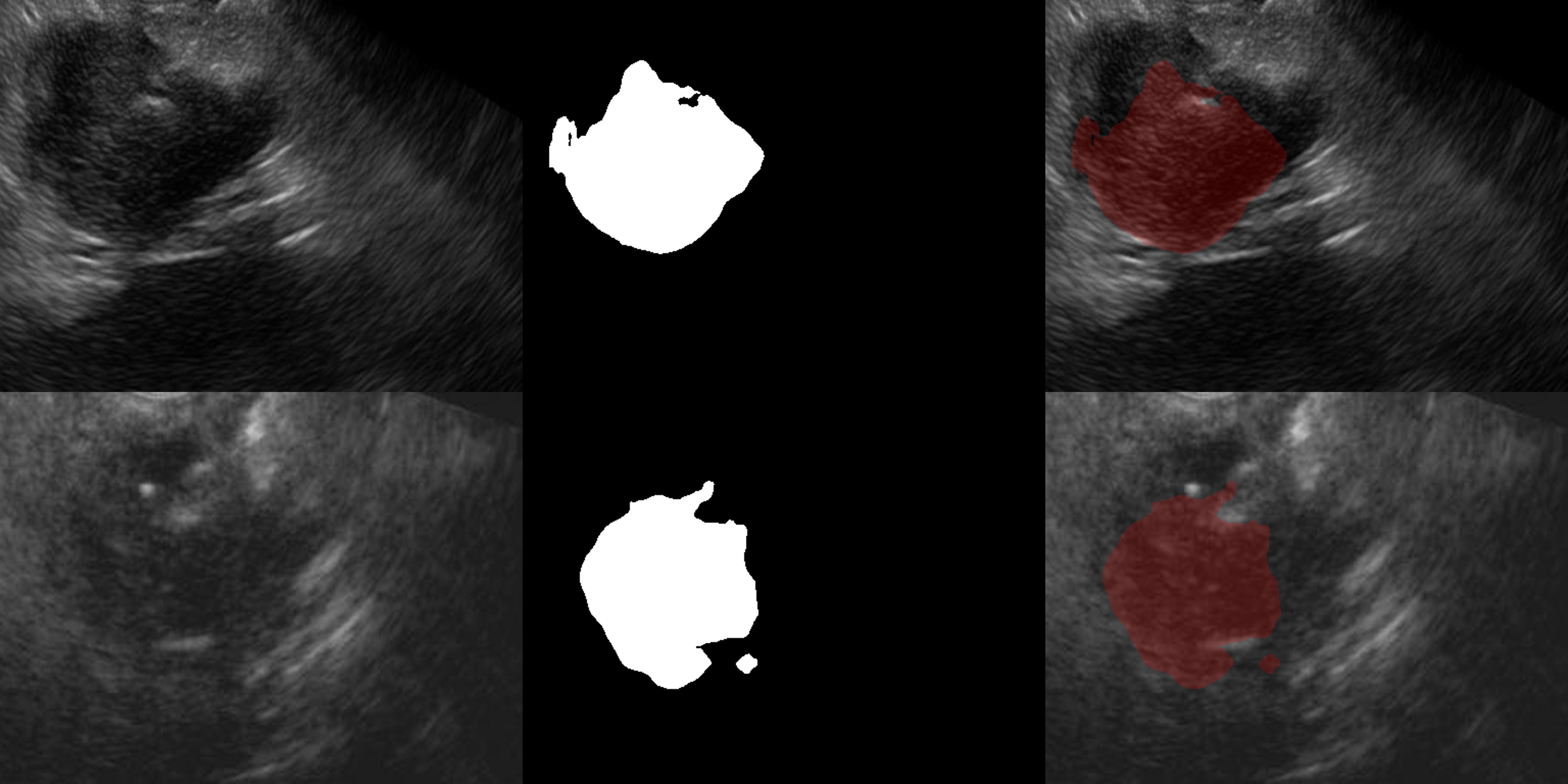}
\caption{Segmentation visualization of pancreatic cancer in 2 patients with lower performance of the DL model. Each row shows (left) input EUS image, (middle) ground-truth mask, (right) model prediction overlay (red). In 1st patient (upper row) the mass has ill-defined boarders (compared with the case shown in Figure 1). The DSC in this image was 0.491. In the 2nd patient (lower row) the mass show subtle difference in the echogenicity compared to the background focally and has ill-defined boarders at places leading to the low Dice score (DSC: 0.413).}
\label{fig:lower-performance}
\end{figure}

Failure analysis: The qualitative analysis of cases with complete failure (n=11) of segmentation with DSC $<$ 0.1 showed a common pattern of lesions being smaller than 1 cm and showing subtle hypoechogenicity. Further cases where DSC $<$ 0.5 were analysed. It was seen that these cases had ill-defined margins (Figure~\ref{fig:lower-performance}).

\section{Discussion}

In this study, we evaluated the performance of a Vision Transformer-based model (HVITBackbone4Seg) for segmenting pancreatic tumors in EUS images. The model achieved a competitive Dice Similarity Coefficient (DSC) of 0.657 and accuracy of 97.5\% on an external test dataset. These results are significant given the inherent challenges of EUS imaging, such as speckle noise, varying echogenicity, and the presence of confounding anatomical structures. The consistent performance metrics across training, validation, and testing phases highlights the robustness of the model.

Our findings align with recent literature emphasized the utility of DL in EUS. For instance, Tang et al.~\cite{ref15} reported a high accuracy of 96\% for pancreatic mass classification, though their segmentation DSC was not explicitly detailed. Similarly, Seo et al.~\cite{ref14} achieved a DSC of 0.81 for pancreatic cancer segmentation using a U-Net architecture. While our DSC is slightly lower, it is important to note that our model was trained on a diverse dataset including both pancreatic cancer and GIST images, adding to the complexity of the task. Furthermore, our use of a ViT-based architecture offers advantages in capturing long-range dependencies compared to traditional CNNs, which is crucial for delineating large or irregular tumors.

The TextSAM-EUS approach proposed by Spiegler et al. further extended segmentation methods by leveraging text prompts to achieve an 82.7\% DSC and 85.3\% normalized surface distance on the public pancreatic cancer dataset that we used for training~\cite{ref17}. The authors tested the model on the split from the same dataset without any external held out testing suggesting overfitting of their model. Additionally, foundation models like SAM demonstrate adaptability but require significant tuning and may struggle with domain-specific challenges such as ultrasound noise and variability. The ViT-based model was trained from an US foundational model, allowing robust segmentation in noisy grayscale EUS images.

There were a few limitations to our study. Manual annotations, while considered gold standard, introduce subjectivity and variability, potentially affecting reproducibility. Limited metadata and demographic information from external test sets restricted detailed subgroup analyses of tumor characteristics. Moreover, a detailed failure cases could not be performed due to limited external dataset details. Additionally, the grayscale nature of the training dataset lacked contrast enhancement, which is known to improve segmentation quality, as demonstrated in Iwasa et al.~\cite{ref13}. The integration of advanced imaging modalities such as contrast-enhanced EUS may further improve segmentation performance and tumor boundary delineation.

Future work can build on this study by adopting multicenter datasets to minimize heterogeneity and improve generalizability. For instance, combining segmentation outputs with diagnostic classification systems, as suggested by Konikoff et al.~\cite{ref16}, could integrate lesion detection and staging into a unified framework. Enhancing segmentation models with post-processing techniques, such as connected-component analysis or filtering, may refine boundary predictions and reduce anomalies.

In conclusion, this study highlighted the robustness and applicability of ViT-based foundational segmentation models for pancreatic tumors on EUS images. By validating on held-out external data, it addressed critical gaps in real-world applicability and generalizability, setting a foundation for future efforts in clinical adoption and refinement of automated EUS workflows.

\bibliographystyle{unsrtnat}
\bibliography{references}

@article{ref1,
  title={Global cancer statistics 2018: GLOBOCAN estimates of incidence and mortality worldwide for 36 cancers in 185 countries},
  author={Bray, Freddie and Ferlay, Jacques and Soerjomataram, Isabelle and Siegel, Rebecca L and Torre, Lindsey A and Jemal, Ahmedin},
  journal={CA: a cancer journal for clinicians},
  volume={68},
  number={6},
  pages={394--424},
  year={2018},
  publisher={Wiley Online Library}
}

@article{ref2,
  title={Sensitivity of CT, MRI, and EUS-FNA/B in the preoperative workup of histologically proven left-sided pancreatic lesions},
  author={Gorris, M and Janssen, QP and Besselink, MG and others},
  journal={Pancreatology},
  volume={22},
  number={1},
  pages={136--141},
  year={2022}
}

@article{ref3,
  title={Challenges in diagnosis of pancreatic cancer},
  author={Zhang, L and Sanagapalli, S and Stoita, A},
  journal={World Journal of Gastroenterology},
  volume={24},
  number={19},
  pages={2047--2060},
  year={2018}
}

@article{ref4,
  title={Impact of endoscopic ultrasonography on diagnosis of pancreatic cancer},
  author={Kitano, M and Yoshida, T and Itonaga, M and Tamura, T and Hatamaru, K and Yamashita, Y},
  journal={Journal of Gastroenterology},
  volume={54},
  number={1},
  pages={19--32},
  year={2019}
}

@article{ref5,
  title={Interobserver Reliability of Endoscopic Ultrasonography: Literature Review},
  author={Yamamiya, A and Irisawa, A and Kashima, K and others},
  journal={Diagnostics (Basel)},
  volume={10},
  number={11},
  pages={953},
  year={2020}
}

@article{ref6,
  title={Deep-learning enabled ultrasound based detection of gallbladder cancer in northern India: a prospective diagnostic study},
  author={Gupta, P and Basu, S and Rana, P and others},
  journal={Lancet Reg Health Southeast Asia},
  volume={24},
  pages={100279},
  year={2023}
}

@article{ref7,
  title={Diagnostic value of deep learning-assisted endoscopic ultrasound for pancreatic tumors: a systematic review and meta-analysis},
  author={Lv, B and Wang, K and Wei, N and Yu, F and Tao, T and Shi, Y},
  journal={Frontiers in Oncology},
  volume={13},
  pages={1191008},
  year={2023}
}

@article{ref8,
  title={Comparison of Vision Transformers and Convolutional Neural Networks in Medical Image Analysis: A Systematic Review},
  author={Takahashi, S and Sakaguchi, Y and Kouno, N and Takasawa, K and Ishizu, K and Akagi, Y and Aoyama, R and Teraya, N and Bolatkan, A and Shinkai, N and Machino, H and Kobayashi, K and Asada, K and Komatsu, M and Kaneko, S and Sugiyama, M and Hamamoto, R},
  journal={Journal of Medical Systems},
  volume={48},
  number={1},
  pages={84},
  year={2024}
}

@inproceedings{ref9,
  title={Endoscopic ultrasound database of the pancreas},
  author={Jaramillo, M and Ruano, J and G{\'o}mez, M and Romero, E},
  booktitle={16th International Symposium on Medical Information Processing and Analysis},
  volume={11583},
  pages={130--135},
  year={2020},
  organization={SPIE}
}

@article{ref10,
  title={Query2: Query over queries for improving gastrointestinal stromal tumour detection in an endoscopic ultrasound},
  author={He, Q and Bano, S and Liu, J and Liu, W and Stoyanov, D and Zuo, S},
  journal={Computers in Biology and Medicine},
  volume={152},
  pages={106424},
  year={2023}
}

@article{ref11,
  title={DSMT-Net: Dual Self-Supervised Multi-Operator Transformation for Multi-Source Endoscopic Ultrasound Diagnosis},
  author={Li, J and Zhang, P and Wang, T and others},
  journal={IEEE Transactions on Medical Imaging},
  volume={43},
  number={1},
  pages={64--75},
  year={2024}
}

@article{ref12,
  title={USFM: A universal ultrasound foundation model generalized to tasks and organs towards label efficient image analysis},
  author={Jiao, J and Zhou, J and Li, X and others},
  journal={Medical Image Analysis},
  volume={96},
  pages={103202},
  year={2024}
}

@article{ref13,
  title={Automatic Segmentation of Pancreatic Tumors Using Deep Learning on a Video Image of Contrast-Enhanced Endoscopic Ultrasound},
  author={Iwasa, Y and Iwashita, T and Takeuchi, Y and others},
  journal={Journal of Clinical Medicine},
  volume={10},
  number={16},
  pages={3589},
  year={2021}
}

@article{ref14,
  title={Semantic Segmentation of Pancreatic Cancer in Endoscopic Ultrasound Images Using Deep Learning Approach},
  author={Seo, K and Lim, JH and Seo, J and others},
  journal={Cancers (Basel)},
  volume={14},
  number={20},
  pages={5111},
  year={2022}
}

@article{ref15,
  title={Endoscopic ultrasound diagnosis system based on deep learning in images capture and segmentation training of solid pancreatic masses},
  author={Tang, A and Gong, P and Fang, N and others},
  journal={Medical Physics},
  volume={50},
  number={7},
  pages={4197--4205},
  year={2023}
}

@article{ref16,
  title={Enhancing detection of various pancreatic lesions on endoscopic ultrasound through artificial intelligence: a basis for computer-aided detection systems},
  author={Konikoff, T and Loebl, N and Benson, AA and others},
  journal={Journal of Gastroenterology and Hepatology},
  volume={40},
  number={1},
  pages={235--240},
  year={2025}
}

@inproceedings{ref17,
  title={Textsam-eus: Text prompt learning for sam to accurately segment pancreatic tumor in endoscopic ultrasound},
  author={Spiegler, P and Koleilat, T and Harirpoush, A and Miller, CS and Rivaz, H and Kersten-Oertel, M and Xiao, Y},
  booktitle={Proceedings of the IEEE/CVF International Conference on Computer Vision},
  pages={948--957},
  year={2025}
}

\end{document}